\pdfoutput=1
\documentclass[11pt]{article}

\usepackage{acl}

\usepackage{times}
\usepackage{latexsym}

\usepackage[T1]{fontenc}
\usepackage[utf8]{inputenc}

\usepackage{microtype}
\usepackage{subfig}
\usepackage{dblfloatfix} % fix for bottom-placement of figure
\usepackage{todonotes}
\usepackage{booktabs}
\usepackage{comment}

\usepackage{amsmath}
\usepackage{amssymb}
\usepackage{arydshln}
\usepackage{booktabs}
\usepackage{tabularx}
\usepackage{tikz}
\usepackage{pgfplots}
\usepackage[normalem]{ulem}
\usepackage{xcolor}
\usepackage{dashbox}%
\usepackage{multirow}
\usepackage{textcomp}
\usepackage{tcolorbox}
\usepackage[ruled, vlined, linesnumbered]{algorithm2e}
\usepackage{mathtools}
\usepackage{soul}

\pgfplotsset{compat=1.13}

\definecolor{plot1}{RGB}{93,147,191}
\definecolor{plot2}{RGB}{233,72,73}
\definecolor{plot3}{RGB}{113,191,110}
\definecolor{plot4}{RGB}{163,73,151}
\definecolor{plot5}{RGB}{230,130,50}
\definecolor{decentgrey}{RGB}{232,232,232}

\usetikzlibrary{calc,fit,positioning,arrows}

\newtcbox{\pattern}{on line,colback=decentgrey,colframe=white,size=fbox,arc=3pt, box align=base, before upper=\strut, top=-2pt, bottom=-2pt, boxrule=0pt}

\pgfdeclarelayer{bg}    % declare background layer
\pgfsetlayers{bg,main}  % set the order of the layers (main is the standard layer)

\newcommand\fmacro{macro-F\textsubscript{1}}
\newcommand\inject{INJECT}
\newcommand\baselineBERT{\texttt{BERT}$\oplus$}
\newcommand\injectBERT{\texttt{BERT}$\otimes$}
\newcommand\baselineROBERTA{\texttt{RoBERTa}$\oplus$}
\newcommand\injectROBERTA{\texttt{RoBERTa}$\otimes$}
\newcommand\baselineELECTRA{\texttt{ELECTRA}$\oplus$}
\newcommand\injectELECTRA{\texttt{ELECTRA}$\otimes$}

\title{Robust Integration of Contextual Information for \\ Cross-Target Stance Detection}

\author{Tilman Beck\thanks{* Equal contribution.} $^{1}$, Andreas Waldis$^{*1,2}$, Iryna Gurevych$^{1}$ \\
$^1$Ubiquitous Knowledge Processing Lab (UKP Lab) \\
  Department of Computer Science and Hessian Center for AI (hessian.AI) \\
  Technical University of Darmstadt \\
 $^2$Information Systems Research Lab \\
 Department of Computer Science, Lucerne University of Applied Sciences and Arts \\
\texttt{\href{http://www.ukp.tu-darmstadt.de/}{www.ukp.tu-darmstadt.de}} \hspace{0.5em} \texttt{\href{http://www.hslu.ch/}{www.hslu.ch}} \\
}

\begin{document}
\maketitle

\begin{abstract}
Stance detection deals with identifying an author's stance towards a target.
%Automatically detecting a stance is critical to understand opinionated discussions on scale.
Most existing stance detection models are limited because they do not consider relevant contextual information which allows for inferring the stance correctly.
Complementary context can be found in knowledge bases but integrating the context into pretrained language models is non-trivial due to the graph structure of standard knowledge bases.
To overcome this, we explore an approach to integrate contextual information as text which allows for integrating contextual information from heterogeneous sources, such as structured knowledge sources and by prompting large language models.
Our approach can outperform competitive baselines on a large and diverse stance detection benchmark in a cross-target setup, i.e. for targets unseen during training.
We demonstrate that it is more robust to noisy context and can regularize for unwanted correlations between labels and target-specific vocabulary.
Finally, it is independent of the pretrained language model in use.\footnote{Data and code at \url{https://github.com/UKPLab/arxiv2022-context-injection-stance}}
%\footnote{Data and code at \url{https://anonymous.4open.science/r/arr-inject-8551/}}

\end{abstract}

\section{Introduction}
\label{sec:intro}

Given a text and a target the text is directed at, stance detection (SD) aims to predict whether the text contains a positive or negative stance towards the target or is not related at all.
We provide an example in \autoref{fig:basic-overview}.
In contrast to formal polls, stance detection (SD) provides a scalable alternative to assess opinions expressed in unstructured texts.
However, in contrast to predicting the polarity of a text (i.e. sentiment analysis), SD is more challenging because it requires establishing the relation towards a target which is rarely mentioned in the text~\citep{augenstein-etal-2016-stance}.

Further, to infer the correct stance, often the text alone is not sufficient and contextual information needs to be taken into account~\citep{dubois2007stance}.
In contrast, most stance classification models are expected to make a correct prediction given the text and target only. 
This can lead to overly relying on label correlations with target-specific vocabulary~\citep{reuver-etal-2021-stance, thorn-jakobsen-etal-2021-spurious}.
In our example (\autoref{fig:basic-overview}), it is challenging to follow the reasoning of the text if the meaning of \textit{school spirit} is left unclear.

\begin{figure}[!htp]
%\footnotesize
\fbox{%
	\begin{minipage}{0.465\textwidth}
	\textbf{Target:} School Uniforms
	
	\textbf{Label:} Pro
	
    \textbf{Text:} Creates a sense of school spirit.
    
    \textbf{Context:} ['school spirit is the enthusiasm and pride felt by the students of a school', 'a strong sense of school spirit is a positive and uplifting influence on the school and its students']
    \end{minipage}
}
    \caption{Example for Stance Detection from the UKP ArgMin dataset~\citep{stab-etal-2018-cross}. The context is not part of the original dataset and was extracted from a large language model via prompting.}
    \label{fig:basic-overview}
\end{figure}

Consequently, providing external knowledge as an additional signal to stance classification has been proposed as a remedy. 
However, lacking a general solution, previous work applies knowledge integration only for a specific text domain like social media~\citep{allaway-etal-2021-adversarial, clark-etal-2021-integrating}.
Nevertheless, SD algorithms are applied on a multitude of different text sources like social media~\citep{ALDAYEL2021102597}, news~\citep{hanselowski-etal-2019-richly} or debating fora~\citep{hasan-ng-2013-stance, chen-etal-2019-seeing} and on diverse targets such as persons~\citep{sobhani-etal-2017-dataset, li-etal-2021-p}, products~\citep{somasundaran-wiebe-2010-recognizing}, or controversial topics~\citep{stab-etal-2018-cross, yohan2021argumentativerelationslogic}, among other things.
In addition, existing approaches~\citep{zhang-etal-2020-enhancing-cross, paul2020argumentative} often depend on the structure of the external knowledge source which is used.
However, a single source of knowledge will likely not suffice for all different scenarios and adapting the model architecture to the structure of a specific knowledge source (e.g. graph-based) limits its applicability.

This work proposes a flexible and robust approach to integrate contextual information by encoding it as text.
It is better aligned to the encoding schema of a pre-trained language model (PLM) and circumvents any dependency on the structure of a particular knowledge source.
It also allows for using any context source that best fits the data's text domain, or mixing contextual information from multiple sources.
In detail, we propose a dual-encoder architecture (INJECT), which encodes the input text and context information separately while facilitating information exchange between both via attention.
We investigate extracting contextual information from various sources using different extraction strategies and evaluate our approach across a benchmark of 16 stance detection datasets exhibiting different characteristics concerning text source, size, and label imbalance.

First, we demonstrate that existing state-of-the-art approaches outperform standard baselines only on the domains they have been tuned for - but perform worse on average.
When integrating context via INJECT, we observe improvements on average and provide an analysis demonstrating the robustness of our approach.
In summary, we make the following contributions:
\begin{itemize}
    \item We show that the performance of existing state-of-the-art approaches does not transfer across a large and diverse benchmark of 16 SD datasets when compared with a standard baseline.
    \item We propose the \inject~architecture to integrate contextual information for cross-target stance detection. Our approach leads to performance improvements across the benchmark and is independent of the underlying pretrained language model.
    \item We provide a comparison of different sources for extracting contextual information and their effectiveness for stance detection. We extract context from structured knowledge bases and by prompting a large pretrained language model. 
    \item An analysis highlights our approach's benefits compared to a more direct integration via appending the context to the input. Our approach regularizes the influence of correlations of target-specific vocabulary and is robust to noisy contexts.
\end{itemize}

\section{Related Work}
\label{sec:rw}

% General Knowledge Injection & Integration
Many tasks in NLP benefit from access to external knowledge such as natural language inference~\citep{chen-etal-2018-neural-natural}, machine translation~\citep{shi-etal-2016-knowledge} or argument mining~\citep{lauscher2022scientia}.
Within the era of PLMs, many approaches rely on extensive pretraining using data from knowledge bases~\citep{peters-etal-2019-knowledge, zhang-etal-2019-ernie} (KB) or supervision from knowledge completion tasks~\citep{wang-etal-2021-k, rozen-etal-2021-teach}.

Early works leveraged sentiment lexicons~\citep{bar-haim-etal-2017-improving} or combinations thereof~\citep{zhang-etal-2020-enhancing-cross} to improve SD classification performance.
Other contextual components like author information~\citep{li-etal-2018-structured, sasaki-etal-2018-predicting, lukasik2019}, dissemination features of social media~\citep{lai2018, veyseh2017} such as retweets or structural discourse elements~\citep{nguyen-litman-2016-context, opitz-frank-2019-dissecting} have been shown to play an important role for stance detection.
Similarly to the aforementioned approaches, focus in SD has shifted towards combining structural KBs and PLMs.
\citet{kawintiranon-singh-2021-knowledge} identify label-relevant tokens and prioritize those during masked language modelling.
This approach risks overfitting on target-specific tokens because stance is often expressed using target-specific terminology - an issue which is particularly problematic for argumentative sentences~\citep{thorn-jakobsen-etal-2021-spurious}. 
\citet{clark-etal-2021-integrating} apply a knowledge infusion method for PLMs by filtering Wikipedia triplets for contextual knowledge.
\citet{popat-etal-2019-stancy} extend BERT by introducing a consistency constraint to learn agreement
between the text and its target.
\citet{jo-etal-2021-classifying} present a variant of BERT pretrained using a variety of supervised tasks resembling logical mechanisms.
\citet{paul2020argumentative} extract relevant concepts from ConceptNet using graph-based ranking methods to improve argument relation classification.
Likewise, \citet{liu-etal-2021-enhancing} use ConceptNet to identify relevant concept-edge pairs and integrate them during training via a graph neural network.
Finally, \citet{hardalov-etal-2021-cross} used label embeddings to improve SD multi-dataset learning and recently showed~\citep{hardalov2022few} that sentiment-based pretraining improves multi-lingual stance detection.

In summary, most existing approaches integrate knowledge through extensive pretraining on knowledge-rich data. 
This does not guarantee improvement of the downstream task they are intended for and requires additional experiments.
Another line of work introduces architectural dependencies on the structure of the knowledge source, thereby limiting their usage to tasks and domains for which the knowledge source is applicable.
In contrast, our approach does not require further pretraining but directly learns to integrate contextual information during supervised training.
The usefulness of the context is, therefore, directly measurable.
Further, our proposed approach integrates context in natural language, thereby decoupling it from the structure of the context source.
It is better aligned with the encoding mechanism of pretrained language models and enables the integration of contextual information from various sources.

\section{Methodology}
\label{sec:methodology}

Our goal is twofold: (1) we aim to integrate contextual information independent of the context source and (2) in a way that is robust to noisy and irrelevant content in the context. 
We propose \inject, a dual encoder approach to integrate contextual sentences using the cross-attention mechanism introduced by \citet{vaswani2017attention}.
The general intuition is that information can flow from input to context and vice versa, thereby regularizing the attention in both encoders.
Thus, the context provides further information to reweigh the prediction importance of individual tokens in the input.

\subsection{Contextual information}

With regards to stance detection, we define \textit{contextual information} (or short \textit{context}) as the sum of all information which, given the text and its target, renders the conclusion of the stance plausible.
The context for each dataset instance is retrieved beforehand and is provided as text to the model.
Formally, we describe context $ \mathbf{c} \in \mathbf{C}$ where $\mathbf{c}$ is a list containing $m$ texts which provide contextual information on the input text $\mathbf{x}$.
See \autoref{fig:basic-overview} for an example ($m=2$).
The length of these texts is upper bounded by the maximum sequence length of the encoder model. 

\begin{figure}
    \centering
    \includegraphics[width=0.49\textwidth]{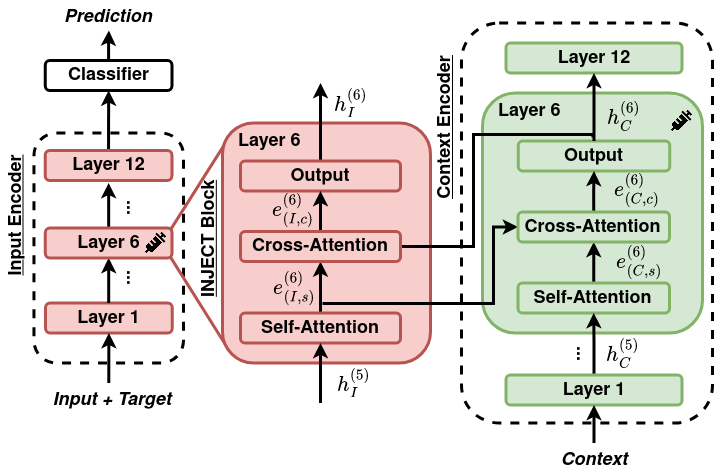}
    \caption{Visualization of the \inject~architecture. It consists of two modules - input encoder and context encoder. The context encoder encodes contextual information, and both encoders are interwoven using an \inject-block based on the cross-attention mechanism.}
    \label{fig:architecture}
\end{figure}

\subsection{Context integration via \inject}
\label{ssec:inject}

\autoref{fig:architecture} provides a high-level visualization of our proposed \inject~architecture.
It consists of two modules: input- and context-encoder.
The input encoder processes input and target $I=(X,T)$ while the context encoder processes the context sentences $C$.
The encoders exchange information using inject blocks ($IB^{(j)}$) which are injected on layer $j$ of both encoders.
$j$ is a hyperparameter which is tuned using the development set of the dataset.
All other layers are standard transformer blocks.
Technically, an $IB$ block is similar to a self-attention block but receives different inputs for key $K$, value $V$ and query $Q$. 
In detail, the inject block of the context-encoder (on layer $j$) receives the output from the self-attention $e^{(j)}_{(I, s)}$ on layer $j$ of the input-encoder as key and value and the output of its own self-attention $e^{(j)}_{(C, s)}$ on layer $j$ as query:
\begin{equation*}
\label{eqn:ib_c}
IB^{(j)}(K{=}e^{(j)}_{(I, s)}, V{=}e^{(j)}_{(I, s)}, Q{=}e^{(j)}_{(C, s)})
\end{equation*}
Afterwards, it is forwarded to get the new hidden state $h_C^{(i)}$ of the context.
Next, we back-inject the context into the input-encoder by feeding $h_C^{(i)}$ as key and value in its inject block:
\begin{equation*}
\label{eqn:ib_i}
IB^{(j)}(K{=}h_C^{(j)}, V{=}h_C^{(j)}, Q{=}e^{(j)}_{(I, s)})
\end{equation*}
Next, the hidden state $h_I^{(i)}$ at layer $j$ of the input encoder is produced by processing the cross-attention output $e^{(j)}_{(I, c)}$.
Finally, we add a classification head to the input encoder, which consists of a pooling layer, dropout and a linear classification layer. 
The parameters of both modules are optimized using the standard cross-entropy loss.

Our architecture is flexible regarding the number of context sentences that can be encoded (parameter $m$).
In case of multiple sentences, we average the cross-attention for all of them.
Due to the dual encoders, \inject~is computationally more efficient compared to context integration via concatenation as we explain in the Appendix~\ref{sec:app:efficiency}.

\subsection{Context integration via concatenation}

An alternative approach would be to append contextual information to the input text such that the model can exploit context directly using the self-attention mechanism.
Technically, this is implemented by separating the input and context using the model-specific separation token (e.g. \texttt{text} + \texttt{[SEP]} + \texttt{context} for BERT)\footnote{In case of two input texts, the context is concatenated to the second input text.}.

We see two major drawbacks of this approach.
First, the integration of irrelevant context will hurt downstream performance due to its direct influence on the attention.
Second, it is limited by the maximum sequence length of the model in use.

\section{Context integration for stance detection}
\paragraph{Task}

In stance detection, given an input text $\mathbf{x} \in \mathbf{X}$ and its corresponding target $\mathbf{t} \in \mathbf{T}$, the goal is to identify the correct label $ \mathbf{y} \in \mathbf{Y}$ from a predefined set of stance descriptions.
We further provide a set of contextual sentences $C$.
The retrieval of $C$ is explained in the next section.

\subsection{Context retrieval}
\label{ssec:kb}

The \inject~model expects the context in natural language form and is therefore flexible with regard to the source of contextual information.
To showcase, we evaluate different context sources that we deem relevant for inferring stance relations: (1) a structured knowledge base which stores knowledge as entity-relationship triplets, (2) a set of causal relations extracted from an encyclopedia, and (3) prompting a large pretrained language model using predefined question templates.
The latter provides an intuitive interface to prompt for relevant sample-specific context, especially in the absence of suitable knowledge bases.

We neither expect these sources to always provide \textit{perfect} contextual information nor to be suitable for all of the heterogeneous stance detection applications (see §\ref{ssec:datasets}).
However, our proposed architecture is designed to be robust, i.e. it utilizes beneficial context and ignores irrelevant information.
In the following, we describe each approach in detail.

\begin{comment}
\begin{figure}
    \centering
    \includegraphics[width=0.49\textwidth]{images/contextsource.png}
    \caption{Different context extraction approaches for SD.}
    \label{tab:kbs}
\end{figure}
\end{comment}

\paragraph{ConceptNet}
Oftentimes, commonsense knowledge is beneficial to correctly infer the stance towards a target and has been shown to complement stance classification~\citep{liu-etal-2021-enhancing}.
Therefore, we use ConceptNet~\citep{conceptnet5.5} which is a directed graph whose nodes are concepts and whose edges are assertions of commonsense about these concepts.
For every edge, ConceptNet provides a textual description of the type of node relationship.
Further, ConceptNet provides a weight factor for every edge computed based on the edge frequency within the ConceptNet training corpus.

Our approach uses the English subset of ConceptNet to get context sentences.
We filter out concepts which are part of English stopwords \footnote{As in NLTK~\citep{bird2006nltk}} and ignore relations without descriptions.
In total, we consider 400k nodes connected through approximately 600k edges. 
To retrieve the context, we use all tokens of the input text to search for string matches within the ConceptNet concepts. 
We consider only paths of length one where the start- and/or end-concept are contained in the input text.
Finally, we sort the paths based on their weight (provided by ConceptNet) and convert every path into a context candidate by joining the descriptions of all its edges, as done in previous work~\cite{lauscher-etal-2020-common}.

\paragraph{CauseNet}
Causal relations, as a more specific example of commonsense knowledge, are often beneficial for understanding opinionated expressions~\citep{sasaki2016stance} but rarely formulated in the text.
To explicate such relations, we investigate CauseNet~\citep{causenet} which is a KB of claimed causal relations extracted from the ClueWeb12 corpus and Wikipedia.
We use the causal relations contained in the high-precision subset\footnote{see \url{https://causenet.org/}} of CauseNet, consisting of 80,223 concepts and  199,806 relations.
We ignore concepts shorter than three characters or consisting of a modal verb (see Appendix~\ref{sec:app:causenet}).
We encode all relations using a sentence encoder~\cite{reimers-gurevych-2019-sentence} using BERT-base-uncased weights. 
For each sample in a dataset, we retrieve the most relevant relations by ranking based on the cosine similarity between the encoded sample and all relations.

\paragraph{Pretrained language model}

Large PLMs can be queried as KBs using natural language prompts~\citep{petroni-etal-2019-language, heinzerling-inui-2021-language}.
We adopt this paradigm and generate context candidates by prompting a PLM to provide more information on either the target, parts of the input or a combination of both.
Precisely, we extract noun-phrases from the input sentence of a length of up to three words using the Stanford CoreNLP tool~\citep{manning-etal-2014-stanford}, ignoring stopwords and filtering noun-phrases which are equal to the target. 
Then, we create prompts using the following templates for single inputs $a$ (e.g. target or noun-phrase) 
\begin{align*}
P_1(a) & = \pattern{define $a$ } \\
P_2(a) & = \pattern{what is the definition of $a$}\\
P_3(a) & = \pattern{explain $a$}
\end{align*}
and combination of inputs $(a,b)$.
\begin{align*}
P_4(a,b) & = \pattern{relation between $a$ and $b$} \\
P_5(a,b) & = \pattern{how is $a$ related to $b$}\\
P_6(a,b) & = \pattern{explain $a$ in terms of $b$}
\end{align*}

The single-input approach is referred to as \texttt{T0pp-NP}, and the second approach as \texttt{T0pp-NP-T}.
We found those prompts to generate the most meaningful contexts across different targets and noun-phrases (see Appendix~\ref{sec:app:prompts} for more details).
The prompts can then be used to generate outputs using any pretrained sequence-to-sequence model.

We make use of T0pp\footnote{See \url{https://huggingface.co/bigscience/T0pp}. We also did experiments using T5 directly but found the outputs to be inferior to T0pp.}~\citep{sanh2022multitask}, which is based on a pretrained encoder-decoder~\citep{raffel2020exploring} and was fine-tuned using multiple diverse prompts generated using a large set of supervised datasets.
We set the output sequence length to 40 words and sort the generated outputs by the length in descending order because we observe T0pp degenerate into producing single words in some cases.
We filter those candidates where more than half of the generated words are repetitions.
Finally, we remove all special tokens from the candidates.
In preliminary experiments, we found using two context sentences ($m=2$) most beneficial.

\section{Experiments}
\label{sec:experiments}

\subsection{Datasets}
\label{ssec:datasets}

\begin{table*}[t]
\centering
    \setlength{\tabcolsep}{3pt}
    \resizebox{1.00\textwidth}{!}{%
    \begin{tabular}{lllll}
    \toprule
    \bf Dataset & \bf Target & \bf Type & \bf Labels & \bf Source \\
    \midrule
    arc~\citep{habernal-etal-2018-argument} &  Headline &   User Post &  unrelated (75\%), disagree (10\%), agree (9\%), discuss (6\%) &  Debates \\
    iac1~\citep{walker-etal-2012-corpus}  & Topic &  Debating Thread & pro (55\%), anti (35\%), other (10\%) &  Debates \\
    perspectrum~\citep{chen-etal-2019-seeing} &  Claim &  Perspective Sent. &                                               support (52\%), undermine (48\%) &       Debates \\
    poldeb~\citep{somasundaran-wiebe-2010-recognizing}                &                  Topic &           Debate Post &                                                     for (56\%), against (44\%) &       Debates \\
    scd~\citep{hasan-ng-2013-stance}                             &  None (Topic) &           Debate Post &                                                     for (60\%), against (40\%) &       Debates \\
    \midrule
    emergent~\citep{ferreira-vlachos-2016-emergent}              &               Headline &               Article &                                   for (48\%), observing (37\%), against (15\%) &          News \\
    fnc1~\citep{pomerleau-2017-FNC}                              &               Headline &               Article &                  unrelated (73\%), discuss (18\%), agree (7\%), disagree (2\%) &          News \\
    snopes~\citep{hanselowski-etal-2019-richly}                  &                  Claim &               Article &                                                    agree (74\%), refute (26\%) &          News \\
    \midrule
    mtsd~\citep{sobhani-etal-2017-dataset}                  &                 Person &                 Tweet &                                      against (42\%), favor (35\%), none (23\%) &  Social Media \\
    rumor~\citep{qazvinian-etal-2011-rumor}                      &                  Topic &                 Tweet &  endorse (35\%), deny (32\%), unrelated (18\%), question (11\%), neutral (4\%) &  Social Media \\
    semeval2016t6~\citep{mohammad-etal-2016-semeval}             &                  Topic &                 Tweet &                                      against (51\%), none (24\%), favor (25\%) &  Social Media \\
    semeval2019t7~\citep{gorrell-etal-2019-semeval}              &  None (Topic) &                 Tweet &                        comment (72\%), support (14\%), query (7\%), deny (7\%) &  Social Media \\
    wtwt~\citep{conforti-etal-2020-will}                    &                  Claim &                 Tweet &                 comment (41\%), unrelated (38\%), support (13\%), refute (8\%) &  Social Media \\
    \midrule
    argmin~\citep{stab-etal-2018-cross}                         &                  Topic &              Sentence &                                   argument against (56\%), argument for (44\%) &       Various \\
    ibmcs~\citep{bar-haim-etal-2017-stance}           &                  Topic &                 Claim &                                                         pro (55\%), con (45\%) &       Various \\
    vast~\citep{allaway-mckeown-2020-zero}                       &                  Topic &             User Post &                                         con (39\%), pro (37\%), neutral (23\%) &       Various \\
    \bottomrule
    \end{tabular}
    }
    \caption{Stance Detection Benchmark datasets and their characteristics (sorted by source, then alphabetically). This table is based on \citet{hardalov-etal-2021-cross}.}
    \label{tab:benchmark}
\end{table*}

We use a SD benchmark~\citep{schiller2021stance, hardalov-etal-2021-cross} which covers 16 datasets altogether in English for research on (cross-domain) stance detection.
We use this benchmark (Table~\ref{tab:benchmark}) because it shows a large diversity regarding text sources, the number of targets, the number of annotated instances, and label imbalance.
Thus, it provides a suitable testbed to evaluate the effectiveness of our context injection approach.
More information about the details of each dataset can be found in the Appendix~\ref{sec:app:datasets}.

\begin{table*}[t!]
 \centering
 \setlength{\tabcolsep}{3pt}
 \resizebox{1\textwidth}{!}{%
 \begin{tabular}{lc|ccccc|ccc|ccccc|ccc}
 
& \fmacro avg. & arc & iac1 & perspectrum & poldeb & scd & emergent & fnc1 & snopes & mtsd & rumor & semeval16 & semeval19 & wtwt & argmin & ibmcs & vast \\

\toprule

\texttt{BERT} & 48.3$\pm$0.7 & 21.5 & 35.6 & 64.6 & \textbf{51.3} & 56.7 & 78.3 & 27.2 & 68.7 & 40.4 & 44.6 & 63.5 & 53.7 & 25.5 & 59.6 & 50.7 & 32.2 \\

\texttt{BERT+Target} & 56.8$\pm$0.8 & 62.5 & 36.3 & \textbf{76.0} &49.8 & 57.9 &78.0 &72.9 & 69.7 &41.2 & 40.5 &64.8 & 53.7 &55.2 &60.3 & 52.0 & 36.1 \\

\midrule

\texttt{STANCY} &  56.2$\pm$0.5 &       62.6 &        36.9 &               75.2 &          50.2 &       57.9 &            78.3 &        \textbf{74.3}$\dagger$ &          \textbf{69.9} &        40.3 &         32.9 &                    64.9 &                 -- &        54.0 &          60.0 &         52.5 &        36.1 \\
\texttt{TGA-Net} & 46.8$\pm$1.4 & 57.2 & 33.9 & 57.5 & 42 & 49.8 & 59.0 & 46.2 & 57.1 & 37.7 & 16.0 & 59.5 & -- & 19.0 & 50.1 & 47.9 & \textbf{62.7}$\dagger$ \\
\texttt{JointCL} & 50.9$\pm$1.8 & 28.6 & 35.8 & 69.6 & 27.2 & 47.1 & \textbf{78.9} & -- & 69.7 & \textbf{55.1}$\dagger$ & 51.5$\dagger$ & \textbf{67.5}$\dagger$ & -- & \textbf{65.1}$\dagger$ & 35.3 & 35.3 & 31.4\\

\midrule

\baselineBERT \texttt{ConceptNet} & 55.7$\pm$0.6 & 61.4 & \textbf{39.3}$\dagger$ & 74.2 & 49.2 & 57.6 & 76.4 & 72.1 & 69.4 & 41.1 & 44.6 & 63.5 & 53.3 & 43.5 & 60.2 & 50.0 & 35.1 \\

\baselineBERT \texttt{CauseNet} & 54.9$\pm$1.3 & 60.6 & 35.0 & 74.4 & 50.0 & 58.0 & 75.0 & 70.9 & 69.2 & 43.2 & 39.1 & 61.1 & 54.3 & 44.5 & 59.4 & 47.3 & 36.0 \\
\baselineBERT \texttt{T0pp-NP} & 55.7$\pm$1.0 & 61.3 & 37.2 & 74.0 & 49.8 & 54.5 & 77.2 & 71.9 & 69.4 & 42.1 & 41.3 & 62.4 & 52.2 & 50.9 & 60.2 & 51.1 & 35.4 \\
\baselineBERT \texttt{T0pp-NP-T} & 56.2$\pm$0.8 & 61.4 & 36.7 & 73.3 & 48.8 & 58.2 & 77.5 & 72.1 & 69.8 & 40.6 & 44.5 & 61.9 & 53.5 & 54.2 & 59.3 & 52.2 & 34.4 \\
\baselineBERT \texttt{All} & 55.5$\pm$1.3 & 61.5 & 38.2 & 74.3 & 49.5 & 56.2 & 75.7 & 70.9 & 68.8 & 43.5 & 42.7 & 62.4 & \textbf{55.3}$\dagger$ & 42.9 & 60.3 & 50.6 & 35.5 \\
\baselineBERT \texttt{Random} & 54.5$\pm$1.1 & 61.3 & 36.3 & 74.5 & 49.7 & 48.3 & 74.8 & 72.1 & 69.6 & 38.4 & 38.2 & 61.8 & 53.8 & 49.6 & 59.0 & 48.2 & 36.1 \\

\midrule

\injectBERT \texttt{ConceptNet} & 57.2$\pm$1.0 & 62.7 & 36.5 & 75.6 & 49.3 & \textbf{58.3} & 77.8 & 73.8 & 69.0 & 41.9 & 47.9 & 65.1 & 54.4 & 52.5 & 60.1 & 53.0 & 37.4$\dagger$ \\
\injectBERT \texttt{CauseNet} & 57.7$\pm$0.9 & 62.9 & 36.9 & 75.5 & 48.9 & 58.0 & 78.1 & 73.6 & 69.3 & 42.4 & 48.1 & 65.7$\dagger$ & 55.1 & 54.8 & 60.7 & \textbf{53.6}$\dagger$ & 39.6$\dagger$ \\
\injectBERT \texttt{T0pp-NP} & 57.5$\pm$1.0 & 62.6  & 37.2 & 75.6 & 48.7 & 57.2 & 77.2 & 73.7 & 69.6 & 41.2 & 49.2$\dagger$ & 65.6$\dagger$ &  55.1 & 55.6 & 60.9 & 52.9 & 37.3$\dagger$ \\
\injectBERT \texttt{T0pp-NP-T} & \textbf{57.8$\pm$1.0} & 62.7 & 37.2 & 75.9 & 49.1 & 57.9 & 78.7 & 74.0$\dagger$ & 69.1 & 41.4 & \textbf{52.2}$\dagger$ & 65.9$\dagger$ & 55.0 & 54.4 & \textbf{61.0} & 53.4$\dagger$ & 37.5$\dagger$ \\
\injectBERT \texttt{All} & 57.2$\pm$0.9 & \textbf{63.0}  & 36.6 & 75.4 & 49.7 & 57.9 & 78.8 & 73.3 & 69.1 & 42.4 & 44.3 & 65.5 & 54.8 & 53.6 & 60.3 & 53.6$\dagger$ & 37.7$\dagger$ \\
\injectBERT \texttt{Random} & 57.3$\pm$1.0 & 62.9 & 36.8 & 75.5 & 49.4 & 57.9 & 78.0 & 73.5 & 69.6 & 41.6 & 45.4 & 65.8 & 54.3 & 54.4 & 60.5 & 53.4$\dagger$ & 37.5$\dagger$ \\

\bottomrule

\end{tabular}
 }
 \caption{Overview of the cross-target results across stance detection benchmark datasets. We highlight best performance per evaluation setting and dataset in bold. Statistically significant differences compared to the best performing baseline without access to context (\texttt{BERT+Target}) are indicated by $\dagger$. Numbers are \fmacro~ scores averaged over ten runs with differently initialized seeds.}
 \label{tab:experiments:results}
\end{table*}

\subsection{Experimental details}

\paragraph{Evaluation}

Our results are evaluated in a cross-target fashion~\citep{augenstein-etal-2016-stance, xu-etal-2018-cross}, i.e. the setup is organized such that instances of a specific target are only contained either in the training, development or test split. 
We point out that our results are not directly comparable to \citet{hardalov-etal-2021-cross} as they perform experiments in a cross-domain fashion, i.e. their goal is to evaluate transfer learning effects by training on \textit{all} but one dataset, which is used for testing.
In contrast, to study the usefulness of context integration, we use \textit{one} dataset per experiment.

\paragraph{Baselines}

We compare \inject~ to the following baselines.
\texttt{BERT} is provided only the input, whereas \texttt{BERT+Target} is provided with both input and target using the model-specific separator token.
(\baselineBERT\texttt{X}) refers to \texttt{BERT+Target} with the retrieved context being appended, where X refers to context sources used (\texttt{ConceptNet}, \texttt{CauseNet},  \texttt{T0pp-NP} and \texttt{T0pp-NP-T}).
We also test a combination of all context sources (\texttt{All}) and integration of random context (\texttt{Random}).
To the best of our knowledge, no prior work has evaluated context integration for cross-target SD on the full benchmark. 
Thus, we compare with \texttt{TGA-Net}~\citep{allaway-mckeown-2020-zero}, \texttt{STANCY}~\citep{popat-etal-2019-stancy}, and \texttt{JointCL}~\citep{liang-etal-2022-jointcl} three state-of-the-art methods for SD which have been proposed for subsets of the benchmark.
% These approaches use varying approaches to overcome the semantic gap between train and test targets. 
\texttt{TGA-Net} uses clustering to identify generalized topic representations.
\texttt{STANCY} applies contrastive learning to learn embeddings where texts supporting a target are closer and opposing texts are more distant to their targets.
\texttt{JointCL} use a prototypical graph for target-aware token representations.
All of them require target information, which is not available for semeval19.
In addition, we found \texttt{JointCL} is not working on fnc1 due to its long input texts.
In these cases, we use the corresponding \texttt{BERT-Target} score for \fmacro~avg. calculation for fair comparison,.

\paragraph{Training setup}
\label{ssec:setup}

We make use of the standard splits given in the benchmark~\citep{hardalov-etal-2021-cross} where possible or create our own (see Appendix~\ref{sec:app:expdetails}).
We use \fmacro~as evaluation metric and average across ten runs with different seeds. 
Performance is measured after the best-performing epoch based on the development set.
To test statistical significance, we use Mann-Whitney U-Test \citep{mann1947test} with $p<0.05$.
For all experiments using BERT, we use the uncased base model~\citep{devlin-etal-2019-bert}.
We use the same set of hyperparameters for all model setups.
For \inject, we use the same model architecture for the input and context encoder.
We tune the layer \textit{j} for context integration (see §\ref{ssec:inject}). 
We tested layers 3, 6, 9 and 12 on the development set of the benchmark. Layer 12 performed the best and was used for all reported results.
More details in the Appendix~\ref{sec:app:expdetails}.

\section{Results}

% shortly summarize main observations: integration of target helps a lot 
In this section, we show with the results on the full benchmark (\autoref{tab:experiments:results}) the effectiveness of \inject~ by providing constant improvements using noisy context for a heterogeneous set of tasks.  

First, we note a large performance boost (+8.5pp) when including information about the target when comparing \texttt{BERT} and \texttt{BERT+Target}. 
While target information improved the performance for individual datasetes~\citep{stab-etal-2018-cross}, we generalize this finding for 14 out of 16 SD datasets.

% baselines work only on the datasets they have been proposed for
The baselines \texttt{STANCY}, \texttt{TGA-Net}, and \texttt{JointCL} mostly show best performance for the datasets they have been proposed for, on average they do not perform on par with the strong \texttt{BERT+Target} baseline.
\texttt{STANCY} performs slightly worse, probably due to the binary contrastive loss and thereby ignoring multi-label information.
\texttt{TGA-Net} clearly underperform all other approaches except for vast. 
Using generalized topic representations transfers to a scenario where the number of targets is relatively high (5634) and only a few examples per target exist (mean 2.4), as for vast.
\texttt{JointCL} performs best on datasets from social media (semeval16, wtwt, or mtsd), but is outperformed by standard baselines for the rest of the tasks. 
Thus, this approach can not generalize to datasets with longer text inputs.
We conclude that existing state-of-the-art approaches for cross-target stance detection work well for the datasets they have been designed for but do not generalize across the diverse set of datasets which exist in SD.

\inject~ outperforms \texttt{BERT+Target} in 13 of 16 cases, while for three datasets (perspectrum, poldeb, snopes) none of the extracted contexts provides benefits, independently of the integration.
On average, all context sources lead to performance improvements in combination with \inject, with \texttt{T0pp-NP-T} being the best.
Combining all context sources underperforms the integration of individual context, most probably due to the average function leading to a perturbation of the context.
Surprisingly, integration of random context slightly outperforms the strong \texttt{BERT+Target} baseline in ten datasets while degrading the direct integration performance, as expected.
We investigate the reasons in our subsequent analysis (§\ref{ssec:analysis}).

\begin{figure*}
    \centering
    \includegraphics[width=1.0\textwidth]{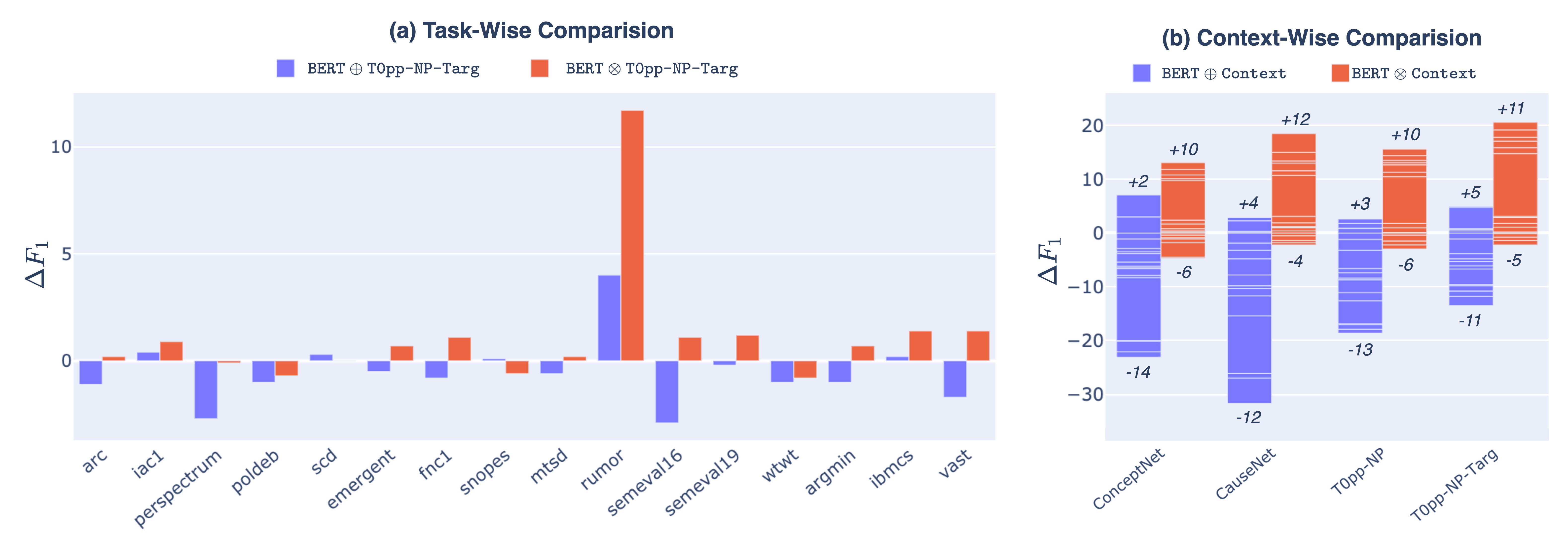}
    \caption{In (a), we compare the relative performance change $\Delta F_1$ of \baselineBERT\texttt{T0pp-NP-T} (blue) or \injectBERT\texttt{T0pp-NP-T} (red) compared to \texttt{BERT+Target} for every task.
    Within (b), we show the aggregated relative performance change of \baselineBERT (blue) and \injectBERT (red) compared to \texttt{BERT+Target} per context source.
    In addition, we count the number of tasks exhibiting performance improvement and deterioration above and below the bars, respectively.}
    \label{tab:inject-vs-direct}
\end{figure*}

\begin{table*}[t]
 %\centering
 \setlength{\tabcolsep}{3pt}
 \resizebox{1.00\textwidth}{!}{%
 \begin{tabular}{lc|ccccc|ccc|ccccc|ccc}
 
  & \fmacro avg. & arc & iac1 & perspectrum & poldeb & scd & emergent & fnc1 & snopes & mtsd & rumor & semeval16 & semeval19 & wtwt & argmin & ibmcs & vast \\

\toprule

\texttt{BERT+Target} & 56.8$\pm$0.8 & 62.5 & 36.3 & \textbf{76.0} & \textbf{49.8} & 57.9 & 78.0 & 72.9 & 69.7 & 41.2 & 40.5 & 64.8 & 53.7 & \textbf{55.2} & 60.3 & 52.0 & 36.1 \\
\baselineBERT \texttt{T0pp-NP-T} & 56.2$\pm$0.8 & 61.4 & 36.7 & 73.3 & 48.8 & \textbf{58.2} & 77.5 & 72.1 & \textbf{69.8} & 40.6 & 44.5 & 61.9 & 53.5 & 54.2 & 59.3 & 52.2 & 34.4 \\
\injectBERT \texttt{T0pp-NP-T} & \textbf{57.8$\pm$1.0} & \textbf{62.7} & \textbf{37.2} & 75.9 & 49.1 & 57.9 & \textbf{78.7} & \textbf{74.0}$\dagger$ & 69.1 & \textbf{41.4} & \textbf{52.2}$\dagger$ & \textbf{65.9}$\dagger$ & \textbf{55.0} & 54.4 & \textbf{61.0} & \textbf{53.4}$\dagger$ & \textbf{37.5}$\dagger$ \\

\midrule

\texttt{RoBERTa+Target} & 61.6$\pm$0.6 &       60.4 &        32.9 &               85.1 &          49.6 &       \textbf{62.3} &            \textbf{79.0} &        \textbf{77.3} &          \textbf{74.9} &        61.2 &         49.9 &                    70.3 &                 57.8 &        \textbf{64.2} &          60.9 &         62.9 &        37.0 \\
\baselineROBERTA \texttt{T0pp-NP-T} &60.8$\pm$0.8 &       61.7 &       \textbf{35.1} &               84.1 &          \textbf{50.6} &       62.1 &            77.8 &        77.0 &          73.9 &        55.2 &         \textbf{51.3} &                    68.2 &                 57.8 &        63.4 &          \textbf{61.6} &         57.9 &        35.6 \\
\injectROBERTA \texttt{T0pp-NP-T} & \textbf{61.9$\pm$0.7} &   \textbf{62.9}$\dagger$ &        33.4 &               \textbf{85.4} &          49.6 &       59.5 &            78.5 &        \textbf{77.3} &          74.6 &        \textbf{64.4} &         51.2 &                    \textbf{70.5} &                 \textbf{58.0} &        62.2 &          61.1 &         \textbf{63.5} &        \textbf{38.5}$\dagger$ \\

\midrule

\texttt{ELECTRA+Target} &62.0$\pm$0.9 &       59.5 &        35.2 &               89.2 &          45.7 &       61.7 &            77.4 &        73.8 &          75.4 &        66.9 &         50.0 &                    \textbf{70.1} &                 55.0 &        63.7 &          60.2 &         71.6 &        36.2 \\
\baselineELECTRA \texttt{T0pp-NP-T} & 61.6$\pm$0.7 &       59.6 &        \textbf{35.5} &               87.5 &          \textbf{47.8} &       \textbf{62.1} &            77.4 &        73.8 &          74.0 &        64.7 &         53.1 &                    67.2 &                 54.1 &        \textbf{65.3} &          60.6 &         68.4 &        35.2 \\
\injectELECTRA \texttt{T0pp-NP-T} & \textbf{63.1$\pm$0.6} &       \textbf{62.5}$\dagger$  &        35.4 &               \textbf{89.3} &          47.4 &       60.4 &            \textbf{78.2} &        \textbf{76.2}$\dagger$ &          \textbf{75.7} &        \textbf{68.9}$\dagger$ &         \textbf{54.7} &                    70.0 &                 \textbf{57.1} &        63.7 &          \textbf{60.7} &         \textbf{71.7} &        \textbf{37.2} \\

\bottomrule

\end{tabular}
 }
 \caption{Comparing context integration using different PLM architectures in a cross-target setup across stance detection benchmark datasets. We highlight best performance per model architecture and dataset in bold. Statistically significant differences compared to the best performing baseline without access to context (\texttt{BERT+Target}) are indicated by $\dagger$. Numbers are \fmacro~ scores averaged over three runs with differently initialized seeds (see Appendix~\ref{sec:app:expdetails} for experimental details.)}
 \label{app:tab:experiments:architecture:results}
\end{table*}

\subsection{Quality of context}
\label{ssec:contextquality}

To evaluate the quality of each context source we used, we looked at the aggregated performance differences with the baseline across each source (\autoref{tab:inject-vs-direct}b).
While \texttt{CauseNet} leads to performance improvements for a maximum number of tasks (12), \texttt{T0pp-NP-T} leads the board with regard to the total sum of absolute improvements across all datasets.
The quality of the context extracted by prompting a PLM also becomes evident when looking at the performance of \baselineBERT.
\texttt{ConceptNet} and \texttt{CauseNet} lead to large performance degradation both in number of tasks and absolute numbers.

\subsection{Generalization across PLMs}
\label{ssec:generalization}

We investigate if the benefits of \inject~ transfer to other PLM architectures by evaluating it in combination with \texttt{RoBERTa}~\citep{roberta2019} and \texttt{ELECTRA}~\citep{electra2020}.
We follow the same experimental protocol as for \texttt{BERT}~\ref{sec:app:expdetails}, but chose only the best-performing context source (\texttt{T0pp-NP-T}) due to the large number of experiments.
The results (\autoref{app:tab:experiments:architecture:results}) confirm the previously observed findings that \inject~ improve the performance on average across this diverse set of stance detection tasks. 
We observe similar improvements as with \texttt{BERT} for both models, with the strongest increase (+1.1pp on average) and the best overall performance for \texttt{ELECTRA}.

\subsection{Further analysis}
\label{ssec:analysis}

As integration with \inject~ outperforms direct integration and even performs more robustly when provided with random context, we analyze the regularization effects the \inject~architecture provides.

\paragraph{Regularization via \inject}

We analyze how \inject~regularize inputs by measure how models rely on target-specific vocabulary.
Such vocabulary is often used to express a stance~\citep{modeling2019wei}, but can lead to spurious correlations~\citep{thorn-jakobsen-etal-2021-spurious}.
Therefore, we identify the top 5\% label-indicative and target-specific tokens and correlate them with the model attributions using vector-norms~\citep{kobayashi-etal-2020-attention} (see Appendix~\ref{sec:app:spurious} for details).
\autoref{tab:analysis-1} shows these correlations for 6 benchmark datasets.
For arc, argmin, and rumor we note a general low to negative correlation of \texttt{BERT+Target}. 
Further, we see \baselineBERT\texttt{T0pp-NP-T} and \injectBERT\texttt{T0pp-NP-T} increasing the correlation - giving more attribution to target and label indicative tokens. 
This behaviour is one reason for the bad performance of \texttt{BERT+Target} for these tasks.
For rumor, we note a correlation increase of 45.5 for \texttt{\inject+T0pp-NP-T}, which leads to a clear performance improvement of 11.7pp (\autoref{tab:experiments:results}).
Thus, \inject~adjusts the low attribution to relevant tokens compared to \texttt{BERT+Target}.
On the other hand, we see \texttt{\inject+T0pp-NP-T} reducing the attribution for ibmcs, mtsd, and wtwt.
Given the better performance of \texttt{\inject+T0pp-NP-T} on ibmcs and mtsd, we conclude that \inject~ can reduce potential spurious correlations in this case. 
For wtwt, \texttt{\inject+T0pp-NP-T} reduces the correlation but has a performance loss of 0.8pp.
Given the niche domain of wtwt (financial mergers and acquisitions on Twitter), identifying relevant context is more challenging using standard context sources.

\begin{table}
  \centering
  \setlength{\tabcolsep}{3pt}
  \resizebox{0.49\textwidth}{!}{%
  \begin{tabular}{lcccccc}
\toprule
        model  &  arc & rumor & argmin  & ibmcs & mtsd & wtwt \\
\midrule
 \texttt{BERT+Target}               & -6.3 & -14.1 &  6.9 &  27.0 &  64.5 & 11.6 \\
 \baselineBERT\texttt{T0pp-NP-T} &  4.5 &  31.0 & 16.2 &  25.7 &  48.9 &  5.3 \\
 \injectBERT\texttt{T0pp-NP-T}   &  8.8 &  31.4 &  9.8 &  22.7 &  33.8 &  6.4 \\
\bottomrule
\end{tabular}
  }
  \caption{Pearson correlation between self-attention and target-label-specific tokens for the baseline model \texttt{BERT+Target} and the context integration approaches (\baselineBERT~ and \injectBERT) using the best performing context source (\texttt{T0pp-NP-T}). A larger correlation indicates stronger attention attribution.}
  \label{tab:analysis-1}
\end{table}

\begin{figure}
    \centering
    \includegraphics[width=0.45\textwidth]{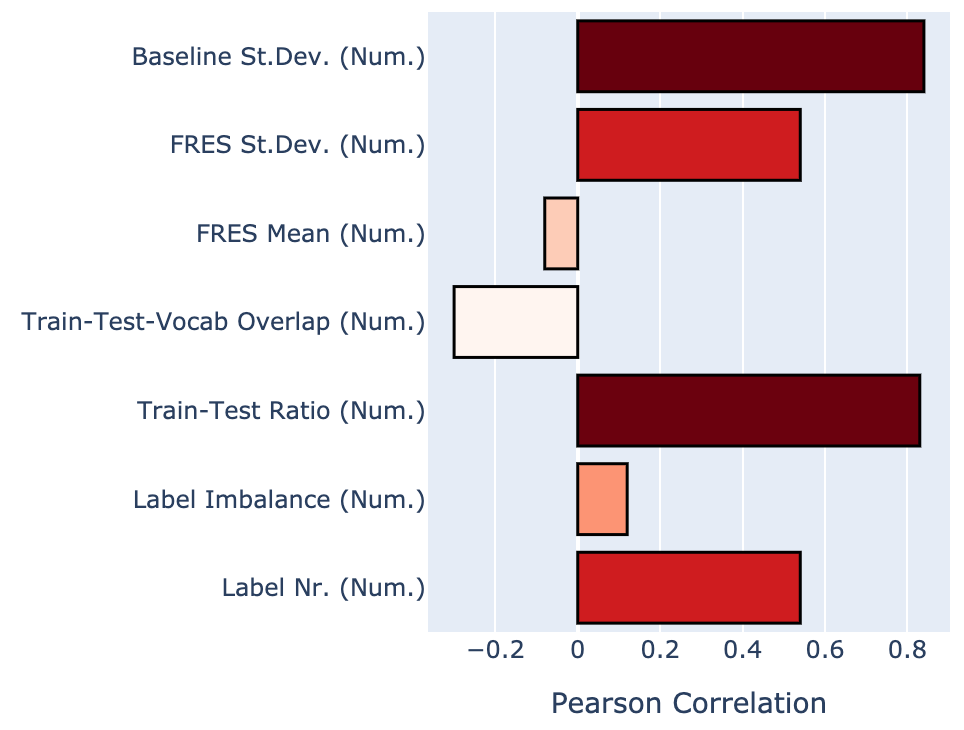}
    \caption{Pearson correlation of various dataset characteristics with performance difference compared to the baseline.}
    \label{fig:pearson-correlations}
\end{figure}

\paragraph{Dataset characteristics}

We investigate which dataset characteristics are indicative of performance improvements using \inject.
Thus, we compute the Pearson correlation of various dataset characteristics, and the performance differences between the baseline and the average of the \inject~ variants.
Details about how we calculate dataset characteristics are provided in the Appendix~\ref{app:sec:dataset-characteristics}.
The results are visualized in Figure~\ref{fig:pearson-correlations}.
Independent of the context source, we observe beneficial improvements using \inject~ if datasets exhibit characteristics leading to performance instabilities.
This is indicated by positive correlations with an increasing number of labels and label imbalance.
Further, we measure text understanding difficulty using the Flesch reading-ease score (FRES) by \citet{flesch1948new}.
Interestingly, \inject~ can better deal with datasets exposing a high variability of FRES within their instances (mean-flesch, std-flesch).
These factors generally contribute to training instabilities where \inject~ is more robust.
This observation is confirmed by the strong positive correlation of the variance across random initializations and \inject~ performance.

\paragraph{Robustness}

We investigate robustness across all benchmark datasets for the \texttt{T0pp-NP-T} context by visualizing the performance differences to the baseline (\texttt{BERT+Target}) in \autoref{tab:inject-vs-direct}a.
In the case of performance improvements, \inject~ consistently outperforms direct context integration.
If there is no improvement for both integration approaches, the performance loss is less pronounced for \inject~with only one exception (snopes).
To substantiate this finding, we contrast both context integration approaches in a scenario with both \textit{ideal} context, i.e. the contextual information is guaranteed to be beneficial in predicting the correct class, and random context.
Our results demonstrate \inject~ successfully leveraging the contextual information while not outperforming direct integration in the case of ideal context.
However, when provided with irrelevant context, \inject~ is closer to a context-free baseline performance.
Details about the experiments are provided in the Appendix~§\ref{sec:app:eval-ideal-context}.
In summary, we conclude that context integration is more robust with regard to noisy context.

\section{Conclusion}
\label{sec:conclusion}

We propose \inject, a dual-encoder approach to integrate contextual information for stance detection based on cross-attention.
While state-of-the-art approaches perform mostly well on the datasets they have been proposed for, we evaluate our approach across a large and diverse benchmark in a cross-target setting and observe improvements compared using three different sources for extracting contextual information.
We show that the context integrated via \inject~improves stance detection and is beneficial for generalization on targets not seen during training.
In future, we plan to explore more sophisticated ways of prompting large pretrained language models for helpful context.

\section*{Ethical Considerations and Limitations}

\paragraph{Quality of the context}
The performance improvement for contextual information injection is bounded by the quality of the context source.
Independently of the source in use, it is possible to introduce additional noise into the training procedure.
While this is a rather generic problem, our proposed architecture seems to be better at filtering noisy context than a direct integration via appending to the input.

\paragraph{Quality of context source}
Most of the existing knowledge bases provide high-quality and curated knowledge. 
In contrast, when prompting a large language model for knowledge, we are also exposed to the risk that we extract the biases (e.g. false facts or stereotypical biases) that the model has learned during pretraining.
In our experiments, we use the T0pp language model where biases have been reported to exist\footnote{More details at \url{https://huggingface.co/bigscience/T0pp?}}.
These biases can potentially influence the prediction performance unintendedly, especially as in many SD datasets, the annotated targets are often controversial.
While investigating such effects is out of scope for this work, we consider such an evaluation inevitable before deploying our proposed model to any data outside (academic) research context. 

\paragraph{Limitations}

As described in §\ref{sec:methodology}, our proposed approach uses two parallel encoder models (input and context). 
It thus requires twice as many parameters as the baseline model we compare to, thereby enforcing additional hardware demands. 
We consider our approach as a proof-of-concept on how to integrate contextual knowledge without amplifying a model's exploitation of spurious correlations. 
We plan to make our architecture more parameter-efficient by investigating more recent approaches for parameter sharing, e.g. with the use of adapters~\citep{houlsbyAdapters2019}.

Moreover, we acknowledge the strong influence of wording in prompts on the output of a language model, as has been reported in the literature~\citep{10.1162/tacl_a_00324, schick-schutze-2021-exploiting}.
We experienced similar effects during preliminary experiments and pointed out that we did not find a one-size-fits-all solution which works equally well across the diverse set of SD benchmark datasets.
Therefore, special care must be taken when extracting contextual information from large language models using prompting.

\section*{Acknowledgements}
We thank Leonardo F. R. Ribeiro, Haritz Puerto, Nico Daheim and the anonymous ARR reviewers for their valuable feedback.
This work has been funded by the German Research Foundation (DFG) as part of the Research Training Group KRITIS No. GRK 2222 and by the German Federal Ministry of Education and Research (BMBF) under the promotional reference 01UP2229B (KoPoCoV) as well as by the Hasler Foundation Grant No. 21024.

% Entries for the entire Anthology, followed by custom entries
\bibliography{anthology,custom}
\bibliographystyle{acl_natbib}

\clearpage

\appendix

\section{Appendix}
\label{sec:app}

\subsection{Experimental Details}
\label{sec:app:expdetails}

\begin{itemize}
    \item All models are trained using five epochs, batch size of 16, a learning rate of 0.00002, a warmup-up ratio of 0.2 with linear scheduling, and AdamW~\citep{adamW2019} as optimizer. The hyperparameters tuned during training are described in the main paper (see §\ref{ssec:setup}).
    \item We use CUDA 11.6, Python v3.8.10, torch v1.10.0, and transformers v4.13.0 as software environment and a mixture of NVIDIA P100, V100, A100, A6000 as GPU hardware.
    \item We load all pretrained language models from \href{https://huggingface.co/}{HuggingFace} model hub. In detail, we use the following model tags: \texttt{bert-base-uncased} for BERT, \texttt{google/electra-base\-discriminator} for ELECTRA, and \texttt{roberta-base} for RoBERTa.
    \item We use the \href{https://captum.ai/}{captum library} (v0.5.0) to calculate the vector-norms for approximating token-attributions~\citep{kobayashi-etal-2020-attention} in §\ref{ssec:analysis}. 
    \item We use the \href{https://www.statsmodels.org/stable/index.html}{statsmodel library}  (v0.13.2) to calculate statistical significant differences using the Bhapkar test \citep{Bhapkar1966ANO} with $p<0.05$.
    \item We use sklearn~\citep{scikit-learn} for computing evaluation metrics (e.g. macro-F1).
    \item We measured the average training runtime of models on the argmin dataset as a reference. \texttt{BERT+Target} and \texttt{BERT+ConceptNet} needed 618 seconds whereas \texttt{\inject} needed 400 seconds.
    \item We use the seeds\texttt{ $[0,1,2,3,4,5,6,7,8,9]$}.
\end{itemize}

\subsection{Datasets}
\label{sec:app:datasets}

We provide details about the individual split proportions for the cross-target evaluation setup in \autoref{tab:dataset:splits:crosstarget}.
For more information on each individual dataset, we refer to \citet{schiller2021stance} and \citet{hardalov-etal-2021-cross}.

\begin{table}[t]
    \centering
    \resizebox{1.00\columnwidth}{!}{%
    \begin{tabular}{l|rrrr} 
    \toprule
    \bf Dataset & \bf Train & \bf Dev &    \bf Test & \bf Total \\
    \midrule
    arc           &  12,382 &  1,851 &   3,559 &  17,792 \\
    argmin        &   6,845 &  1,568 &   2,726 &  11,139 \\
    emergent      &   1,638 &    433 &     524 &   2,595 \\
    fnc1          &  42,476 &  7,496 &  25,413 &  75,385 \\
    iac1*          &   4,221 &    453 &     923 &   5,597 \\
    ibmcs         &     935 &    104 &   1,355 &   2,394 \\
    mtsd         &   6,227  &   1,317 &   1,366 &   8,910 \\
    perspectrum   &   6,978 &  2,071 &   2,773 &  11,822 \\
    poldeb        &   4,753 &  1,151 &   1,230 &   7,134 \\
    rumor*        &   6,093 &    299 &     505 &   7,106 (10,237) \\
    scd           &   3,251 &    624 &     964 &   4,839 \\
    semeval2016t6 &   2,497 &    417 &   1,249 &   4,163 \\
    semeval2019t7* &   5,205 &  1,478 &   1,756 &   8,439 (8,529) \\
    snopes        &  14,416 &  1,868 &   3,154 &  19,438 \\
    vast          &  13,477 &  2,062 &   3,006 &  18,545 \\
    wtwt          &  25,193 &  7,897 &  18,194 &  51,284 \\
    \bottomrule
    % \midrule        
    % Total         & 154,228 & 30,547 & 68,495  & 253,270 \\ 
    % \bottomrule

    \end{tabular}
    }
    \caption{Number of examples per data split for the cross-target evaluation setting. For datasets marked with \textsuperscript{*}, not all tweets could be downloaded or we discovered empty instances which we excluded (in comparison to the numbers provided by \citet{hardalov-etal-2021-cross}); for mtsd, we received the full dataset by the original authors; the original number of tweets is in parentheses.}
    \label{tab:dataset:splits:crosstarget}
\end{table}

\subsection{Evaluation with ideal context}
\label{sec:app:eval-ideal-context}

To evaluate our goal of robust integration of contextual information using \inject, we contrast both context integration approaches in a scenario with both \textit{ideal} context, i.e. the contextual information is guaranteed to be beneficial in predicting the correct class, and random context.
To showcase, we use the e-SNLI~\citep{esnli} corpus for natural language inference and the Snopes~\citep{hanselowski-etal-2019-richly} corpus for claim verification.
We use the provided explanations (e-SNLI, m=1) and evidences (Snopes, m=10) as ideal context, respectively.
As random (but syntactically correct) context, we randomly extract sentences from the Gutenberg corpus\footnote{\url{http://www.gutenberg.org/}} included in NLTK~\citep{bird2006nltk}.
Table~\ref{tab:method:eval} compares a BERT baseline without context, BERT with context integration via concatenation (\baselineBERT), and integration via \inject~(\injectBERT).

The results demonstrate \inject~ successfully leveraging the contextual information while not outperforming direct integration in the case of ideal context.
However, when provided with irrelevant context, \inject~ is closer to the context-free baseline performance.

\begin{table}
\tiny
  \centering
  \setlength{\tabcolsep}{3pt}
  \resizebox{0.49\textwidth}{!}{%
  \begin{tabular}{lcccc}
\toprule
         {} & e-SNLI (Ideal) & e-SNLI (Random) & Snopes (Ideal) & Snopes (Random) \\
  \midrule
  \texttt{BERT} & 90.33 & 90.33 & 51.8 & 51.8 \\
  \baselineBERT & 98.70 & 90.08 & 78.0 & 49.7 \\
  \injectBERT   & 98.35 & 90.52 & 75.8 & 51.1 \\
\bottomrule
\end{tabular}
  }
  \caption{Comparison of context integration via concatenation (\baselineBERT) and \inject~ (\injectBERT) on e-SNLI and Snopes. Original dataset splits are used. Scores are macro-F1 averaged across three seeds.}
  \label{tab:method:eval}
\end{table}

\subsection{Identification of target-specific label correlations}
\label{sec:app:spurious}

\begin{figure*}
  \centering
  \includegraphics[width=0.95\textwidth]{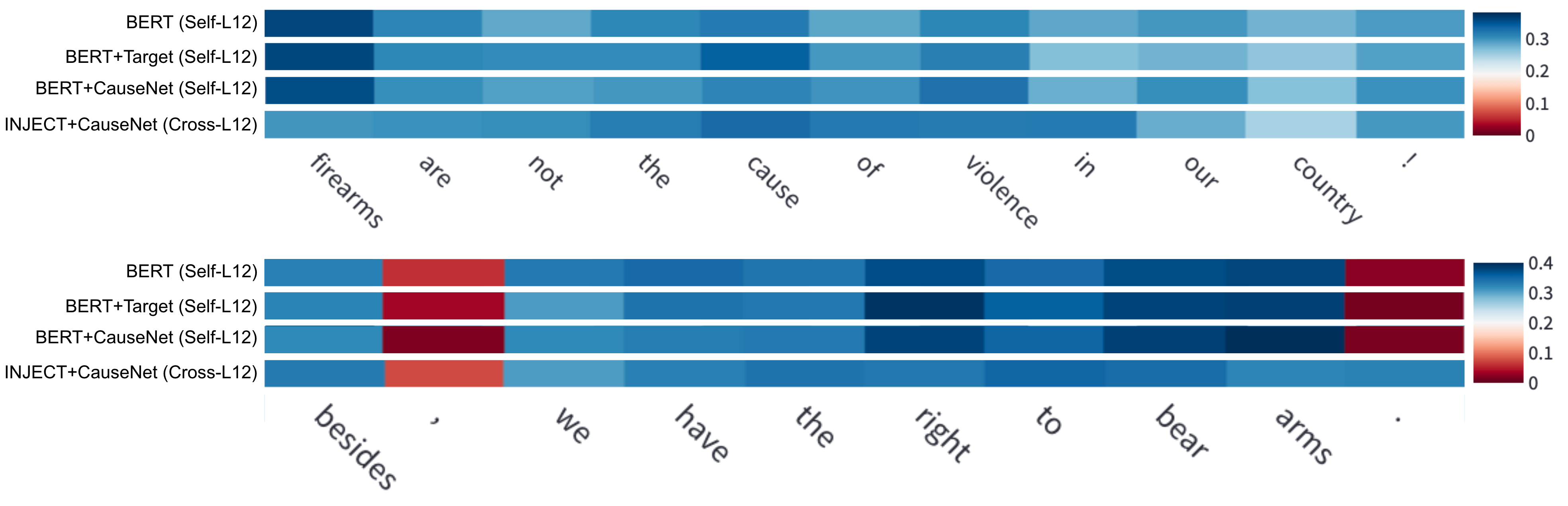}
  \caption{Two examples of the argmin dataset. The first is an argument against gun control, while the second supports it. It shows the token-level attribution for \texttt{BERT}, \texttt{BERT+Target}, \texttt{BERT+CauseNet}, and \texttt{\inject+CauseNet}.}
  \label{fig:examples}
\end{figure*}

We examine internal processes in the model architecture by analyzing how relevant a token is compared to how much a model attributes to the token. 
In detail, we calculate for the 5\% most relevant tokens for target and label the correlation of this relevance and the model attribution on them.

\paragraph{Token Relevance}
We consider the probability of a token to appear in combination with a label $\mathbf{l}$ and target $\mathbf{t}$. 
A higher probability indicates that a token is more likely to occur within a label-target combination. 

In detail, we first calculate the relevance as the maximum log-odds-ratio $r_{(w, (l_i, t_j))}$ \citep{kawintiranon-singh-2021-knowledge} over all possible combinations of labels $L=\{l_1, ..., l_n\}$ and targets $T=\{t_1,...,t_k\}$ for a given token $w$.
We define $o_{(w, (l, t))}$ (\autoref{eqn:odds}) as the probability of token $t$ appearing in combination with label $l$ and target $t$, with $c(w, (l, t))$ denoting the counts of $w$ in texts with label $l$ and target $t$. 
%E.g. in case of the target, this is the odds of observing a token in a specific target $p$ and not in the others. 
Next, we calculate the maximum log odds-ratio $r_{(t, (L, T))}$ as in \autoref{eqn:odds-ratio}.
This tells us how specific a token $w$ is at max. for a label-target combination. 

\begin{equation}
\label{eqn:odds}
    o_{(w, (l_i, t_j))}=\frac{c(w, (l_i, t_j))}{c(\neg w, (l_i, t_j))}
\end{equation}
\begin{equation}
\label{eqn:odds-ratio}
    r_{(w, (L, T))} = \mathop{max}_{(l_i, t_j) \in L \times T} log(\frac{o_{(w, (l_i, t_j))}}{o_{(w, \neg(l_i, t_j))}})
\end{equation}

\paragraph{Token Attributions}
To approximate a token's attribution, we calculate the vector-norms \cite{kobayashi-etal-2020-attention} for the output of the 12th layer. 

We provide anecdotal examples in \autoref{fig:examples} along with their token-level attribution of the 12th layer from (\texttt{BERT}, \texttt{BERT+Target}, \texttt{BERT+CauseNet}) and \texttt{\inject+CauseNet}.
For the first three, we use the self-attention and for the latter one the cross-attention.
In the first example, \texttt{\inject+CauseNet} made the right prediction while all \texttt{BERT}-based models failed and vice-versa for the second one. 
In both examples, we see lower attribution for target-specific terms like \textit{firearms} or \textit{arms} and higher attribution for terms with general use like \textit{besides}, \textit{cause}, or \textit{to}.
\texttt{\inject+CauseNet} makes the correct prediction while \texttt{BERT+Target} failed due to its high attribution to \textit{firearms} - an example of a spurious correlation.
However, in some cases this can also lead to erroneous predictions as in the second example where \texttt{\inject+CauseNet} gives less importance to the specific - and in this case important - tokens of the sentences (\textit{right to bear arms}).

\subsection{Dataset Characteristics}
\label{app:sec:dataset-characteristics}

\begin{table*}[t]
 \centering
 \setlength{\tabcolsep}{3pt}
 \resizebox{1\textwidth}{!}{%
 \begin{tabular}{l|ccccc|ccc|ccccc|ccc}
 
  & arc & iac1 & perspectrum & poldeb & scd & emergent & fnc1 & snopes & mtsd & rumor & semeval16 & semeval19 & wtwt & argmin & ibmcs & vast \\

\toprule

Number of Labels & 4 & 3 & 2 & 2 & 2 & 3 & 4 & 2 & 3 & 5 & 3 & 4 & 4 & 2 & 2 & 2 \\ 
Label Imbalance & 1,16 & 0,55 & 0,03 & 0,11 & 0,20 & 0,41 & 1,13 & 0,48 & 0,23 & 0,61 & 0,37 & 1,09 & 0,58 & 0,11 & 0,11 & 0,22 \\
Train-Test-ratio  & 3.48 & 4.57 & 2.52 & 3.86 & 3.37 & 3.13 & 1.67 & 4.57 & 4.56 & 12.48 & 2 & 2.96 & 1.38 & 2.51 & 0.69 & 4.48 \\ % 0.85
Train-Test-Vocabulary Overlap & 16461 & 16939 & 2875 & 7553 & 5171 & 861 & 11244 & 8092 & 4876 & 818 & 2666 & 3248 & 7836 & 4454 & 1361 & 6271\\ % -0.32
FRES Mean & 63.07 & 70.27 & 53.12 & 65.11 & 70.70 & 66.37 & 61.44 & 61.36 & 71.43 & 58.08 & 67.43 & 58.62 & 48.41 & 51.73 & 39.94 & 63.19 \\
FRES St.Dev. & 14.2 & 13.6 & 29.6 & 37.4 & 29.5 & 21.5 & 10.3 & 26.3 & 17.5 & 57.7 & 22.2 & 49.3 & 26.8 & 22.8 & 29.4 & 14.1 \\ % 0.54
Baseline St.Dev. & 0.8 & 2.9 & 0.8 & 2.8 & 1.7 & 1.5 & 1.3 & 0.8 & 2.1 & 9.8 & 0.6 & 2.6 & 4.1 & 1.4 & 1.5 & 1\\ % -0.81

\bottomrule

\end{tabular}
 }
 \caption{Overview of the dataset-characteristic for each dataset. }
 \label{app:tab:datasetcharacteristics}
\end{table*}

In \autoref{app:tab:datasetcharacteristics}, we provide relevant dataset characteristics for each dataset in the stance detection benchmark.
To compute label imbalance, we first calculate the mean and standard deviation of the number of instances per label.
The label imbalance is then defined as the division of the standard deviation by the mean.

\subsection{Knowledge}
\label{sec:app:knowledge}

\begin{table*}[t]
  \centering
  \setlength{\tabcolsep}{3pt}
  \resizebox{1\textwidth}{!}{%
  \begin{tabular}{lcccccccccccccccc}
    \toprule
    Knowledge Source  &  arc &  iac1 & perspectrum &   poldeb &     scd &  emergent &    fnc1 &   snopes &    mtsd &    rumor & semeval16 & semeval19 &    wtwt &   argmin &    ibmcs &    vast \\
    \midrule
     \texttt{ConceptNet} & 5.1 & 5.1 & 5.5 & 5.1 & 5.2 & 5.5 & 5.1 & 5.5 & 5.6 & 6.0 & 5.5 & 5.6 & 5.6 & 5.3 & 5.3 & 5.1 \\
     \texttt{CauseNet} & 91.2 & 112.1 & 20.5 & 78.4 & 69.8 & 34.1 & 137.6 & 40.3 & 56.4 & 50.6 & 52.1 & 47.0 & 43.0 & 36.4 & 23.7 & 89.5 \\
     \texttt{T0pp-NP} & 13.1 & 13.1 & 13.0 & 12.9 & 12.5 & 13.3 & 14.1 & 13.6 & 12.9 & 13.1 & 12.4 & 13.0 & 12.5 & 12.5 & 13.5 & 13.1 \\
     \texttt{T0pp-NP-T} & 9.9 & 12.7 & 10.5 & 12.1 & 11.9 & 11.6 & 16.7 & 11.9 & 14.0 & 13.6 & 12.7 & 9.4 & 11.1 & 12.7 & 11.7 & 12.2 \\
    \bottomrule
    \end{tabular}
    }
  \caption{Average length for each combination of knowledge extraction method and dataset.}
  \label{tab:contextlength}
\end{table*}

The information about the average length of the retrieved contextual knowledge is given in \autoref{tab:contextlength}.
We observe substantially longer paragraphs extracted from CauseNet which is not surprising as CauseNet consists of passages extracted from Wikipedia.

\subsubsection{CauseNet}
\label{sec:app:causenet}

We ignore concepts which are shorter than 3 characters or consist of one of the following modal verbs ("must", "shall", "will", "should", "would", "can", "could", "may", "might").

\subsubsection{Prompts}
\label{sec:app:prompts}

We manually evaluated the following prompts for both single and combination inputs.
As reported in related work~\citep{10.1162/tacl_a_00324, schick-schutze-2021-exploiting}, the generated text is sensible to wording and punctuation in the prompt.
We made similar experiences and removed all punctuation at the end of the prompt to prevent the model from generating outputs of short length.

\begin{table}
\tiny
\centering
    \resizebox{0.4\textwidth}{!}{%
    \begin{tabular}{lc} 
    \toprule
    \bf Prompt & \bf Usage \\
    \midrule
    define $a$ & \checkmark \\
    what is $a$ &  \\
    describe $a$ & \\
    what is the definition of $a$ & \checkmark \\
    explain $a$ & \checkmark \\
    relation between $a$ and $b$ & \checkmark \\
    how is $a$ related to $b$ & \checkmark \\
    explain $a$ in terms of $b$ & \checkmark \\
    \bottomrule
    \end{tabular}
    }
    \caption{Prompts which have been evaluated for generating contextual knowledge for stance detection.}
    \label{tab:prompts}
\end{table}

\subsection{On Efficiency of \inject}
\label{sec:app:efficiency}

From an efficiency point-of-view, \inject processes a text and corresponding contexts more efficiently than via SEP integration. This is because there is no self-attention over \textit{input} and \textit{context} jointly where the attention dimension is $d_{sep} = (\text{len}(input)+\text{len}(context))\times(\text{len}(input)+\text{len}(context))$. For \inject, in contrast, input and context are processed in separate encoders with attention dimensions $d_{input} = \text{len}(input)\times\text{len}(input)$ and $d_{context} = \text{len}(context)\times\text{len}(context)$ on every layer. Just in the \inject layer, there are two additional attention blocks with dimensions $d_{\textit{cross context}} = \text{len}(input)\times\text{len}(context)$ and $d_{\textit{cross input}} = \text{len}(context)\times\text{len}(input)$.

\end{document}